\documentclass{article}
\usepackage{ijcai24}

\usepackage{times}
\usepackage{soul}
\usepackage{url}
\usepackage[hidelinks]{hyperref}
\usepackage[utf8]{inputenc}
\usepackage[small]{caption}
\usepackage{graphicx}
\usepackage{amsmath}
\usepackage{amsthm}
\usepackage{booktabs}
\usepackage{algorithm}
\usepackage{algorithmic}
\usepackage[switch]{lineno}
\usepackage{multirow}
\usepackage{pifont}
\newcommand{\cmark}{\ding{51}}%
\newcommand{\xmark}{\ding{55}}%

\title{Towards Scale-Aware Full Surround Monodepth with Transformers}

\author{
Yuchen Yang$^1$
\and
Xinyi Wang$^2$
\and
Dong Li$^1$
\and
Lu Tian$^1$
\and
Ashish Sirasao$^1$
\And
Xun Yang$^2$
\\
\affiliations
$^1$AMD\\
$^2$USTC\\
\emails
ethany@amd.com,
wxy1@mail.ustc.edu.cn,
\{dongl, lu.tian, ashish.sirasao\}@amd.com,
xyang21@ustc.edu.cn
}

\begin{document}

\maketitle

\begin{abstract}
Full surround monodepth (FSM) methods can learn from multiple camera views simultaneously in a self-supervised manner to predict the scale-aware depth, which is more practical for real-world applications in contrast to scale-ambiguous depth from a standalone monocular camera. In this work, we focus on enhancing the scale-awareness of FSM methods for depth estimation. To this end, we propose to improve FSM from two perspectives: depth network structure optimization and training pipeline optimization. 
First, we construct a transformer-based depth network with neighbor-enhanced cross-view attention (NCA). 
The cross-attention modules can better aggregate the cross-view context in both global and neighboring views.
Second, we formulate a transformer-based feature matching scheme with progressive training to improve the structure-from-motion (SfM) pipeline. That allows us to learn scale-awareness with sufficient matches and further facilitate network convergence by removing mismatches based on SfM loss. 
Experiments demonstrate that the resulting Scale-aware full surround monodepth (SA-FSM) method largely improves the scale-aware depth predictions without median-scaling at the test time, and performs favorably against the state-of-the-art FSM methods, e.g., surpassing SurroundDepth by 3.8\% in terms of accuracy at $\delta<1.25$ on the DDAD benchmark.
\end{abstract}

\section{Introduction}

Accurate perception of the 3D world is crucial for the success of autonomous driving and robotic applications. While traditional 3D sensors like LiDAR can provide this information, monocular cameras offer a cost-efficient alternative solution that is capable of dense depth prediction and therefore providing richer context information. This information is useful for upper-level tasks such as 3D detection and segmentation.
Although depth prediction from monocular cameras is an ill-posed problem, several companies have deployed “Vision-based LiDAR” systems using monocular setups, achieving sensor redundancy or the replacement of radar and/or laser-based ranging techniques. 

Monocular depth prediction can be learned in self-supervised ways and has been a popular research topic due to its low-cost characteristic during both training and inferencing. This self-supervised learning is typically achieved by leveraging the ego-motion in video sequences, such as using a pose network that receives temporal adjacent frames and predicts pose changes. However, since the relative position changes among 2D frames are measured without an absolute real-world scale, an auxiliary re-scaling step is required to project the predicted relative depth to match the true distance using the ground truth. This step is referred to median-scaling. Although evaluation with median-scaling may appear to yield accuracy that is comparable to supervised learning, it may not be practical for real-world applications because obtaining groundtruth scale may not be straightforward during inferencing.
One way to learn the absolute scale depth is by leveraging a binocular setup, as the binocular calibration step finds corresponding points across cameras on distance-aware calibration patterns such as chessboard patterns.
However, it may not be an ideal solution since it requires highly overlapped camera views and rectification pre-steps for the algorithm to be trained, and it doubles the number of cameras while providing a limited field-of-view depth.
Another way to obtain the real-world scale is to use geometry with additional information such as camera height or the size or distance of objects in the scene. However, acquiring such information can be challenging and prone to errors, making it difficult to achieve accurate scale-aware depth estimation.
Recent advancements in Full surround monodepth (FSM) have introduced promising approaches for capturing a 360-degree field-of-view by placing multiple cameras with known camera matrices. FSM reduces the cost for calibration and setup as each camera covers a large field of view and has only a small overlap with neighbor cameras. With known camera matrices and self-supervised learning, FSM can learn scale-aware depth, without requiring an additional median-scaling step. 
While several works have been published \cite{guizilini2022full,xu2022multi,kim2022self,wei2022surrounddepth} using the multi-camera setting, further enhancements are possible.

In this paper, we propose a transformer-based depth network for predicting 360-degree depth to further enhance the scale-awareness for FSM depth prediction.
Our key innovation involves introducing a novel Neighbor-enhanced Cross-View attention (NCA) module to aggregate the context in both global and neighboring views. 
Different from prior work that only considers global information (e.g., CVT in Surrounddepth), our NCA module is designed to facilitate the exchange of cross-view context among strongly correlated regions from neighboring views. 


Moreover, we optimize the structure-from-motion pipeline with two-round progressive training. In the first round, we collect all the matching points to create a pseudo groundtruth for supervise training. Using sufficient matches can help learn scale-awareness and quickly derive a good initialization. In the second round, we adopt a simple filtering strategy based on SfM loss to enable only valid points for training. Removing mismatches can help facilitate the network convergence in this phase. Such progressive training strategy is simple but effective to boost scale-aware depth prediction. 

We summarize our contributions as follows:
(1) We propose a transformer-based depth network for FSM depth estimation that aggregates the cross-view context in both neighboring and global views.
(2) We propose an optimized training pipeline that adopts transformer-based feature matching with progressive training to boost scale-awareness. 
(3) Our proposed approach, named SA-FSM, improves scale-aware depth prediction and outperforms previous FSM methods on the DDAD benchmark.

\section{Related Works}

\subsection{Monocular Depth Estimation}

Previous works that utilize monocular cameras for unsupervised depth learning typically leverage clues from stereo pairs and/or video sequences \cite{godard2017unsupervised,zhou2017unsupervised,godard2019digging,guizilini20203d}. These approaches benefit from view synthesis among frames, either in adjacent views or in temporal sequence.
Stereo pairs enable true scale depth learning through stereo rectification and calibration, but the extensive overlapping area assumption they rely on is cost-inefficient.
Methods learning from video sequences alone, though provide flexibility in setting, require median-scaling during evaluation.
Recent work proposed a teacher-student learning scheme with pseudo label, and estimates true scale by aligning relative and actual camera heights \cite{petrovai2022exploiting}.
Though investigated, learning the true scale from video sequence alone remains an uneasy task for self-supervised monocular depth in training. 
In FSM, models can encode the real-world scale into the depth map estimation with only small overlapped regions among cameras by leveraging the extrinsic matrices and spatial constraints in the overlapping area. 

With the success of transformers being applied to vision tasks \cite{dosovitskiy2020image}, several transformer-based methods emerged for supervised monocular depth estimation \cite{ranftl2021vision,agarwal2022depthformer,yang2021transformer}. 
For unsupervised setting, MonoViT \cite{zhao2022monovit} adds plain convolution layers alongside with transformer to form a joint layer to capture local and global information.
DEST \cite{yang2022depth} adopted a variant of transformer - Segformer \cite{xie2021segformer} and made further efficiency improvements including simplified transformer blocks, feature sharing, and progressive decoder.
Unlike previous transformer-based networks used in monocular depth estimation task, our proposed FSM depth network handles multiple camera views and conducts context exchange among the views within the network with transformer blocks. 

\subsection{Full Surround Monodepth}
The pioneer work of full surround monodepth (FSM) challenge \cite{guizilini2022full} proposed to learn the scale-aware depth by using spatial-temporal consistency with photometric loss, which operates on adjacent and temporal frames. Additionally, they added a pose constraint that bounds the camera pose for each camera, and applied photometric masks to non-overlapping and self-occluded areas.
MCDP \cite{xu2022multi} treated the final depths as a weighted combination of depth basis. They proposed a weight network that uses a concatenation of warped adjacent features and target features to refine the depth basis produced by the depth network. 
VFF \cite{kim2022self} constructed a 3D volumetric feature in voxel coordinates with MLPs. Their method is flexible in projecting 3D points to novel views with varying focal lengths.
Surrounddepth \cite{wei2022surrounddepth} proposed a cross-view attention transformer for global feature aggregation, a joint-pose prediction that unifies the pose from multiple cameras, and additional pre-training with structure-from-motion (SfM). 
EGA-Depth \cite{shi2023ega} proposed a model for scale ambiguous depth prediction with guided attention across camera view and achieved high efficiency and accuracy. 
R3D3 \cite{schmied2023r3d3} proposed to use feature correlation and bundle adjustment for depth and pose refinement. Their method requires additional tuning with synthetic dataset and iterative inference based on sequence of frames.

Our method differs from previous FSM methods in the following aspects:
(1) We aim to improve accuracy for scale-aware depth prediction through network and training pipeline improvements.
(2) We uniquely aggregate the strongly correlated context from adjacent camera through a novel NCA module. This design allows the depth network to effectively leverage context information and outperforms previous designs which aggregates context from all camera views.
(3) Compared with previous SfM pipeline \cite{wei2022surrounddepth}, we offer an enhanced progressive approach that utilizes a stronger descriptor and a tailored filtering scheme to provide accurate real-world scale guidance during training.

\section{Method}

\begin{figure*}
\begin{center}
\includegraphics[width=0.85\textwidth]{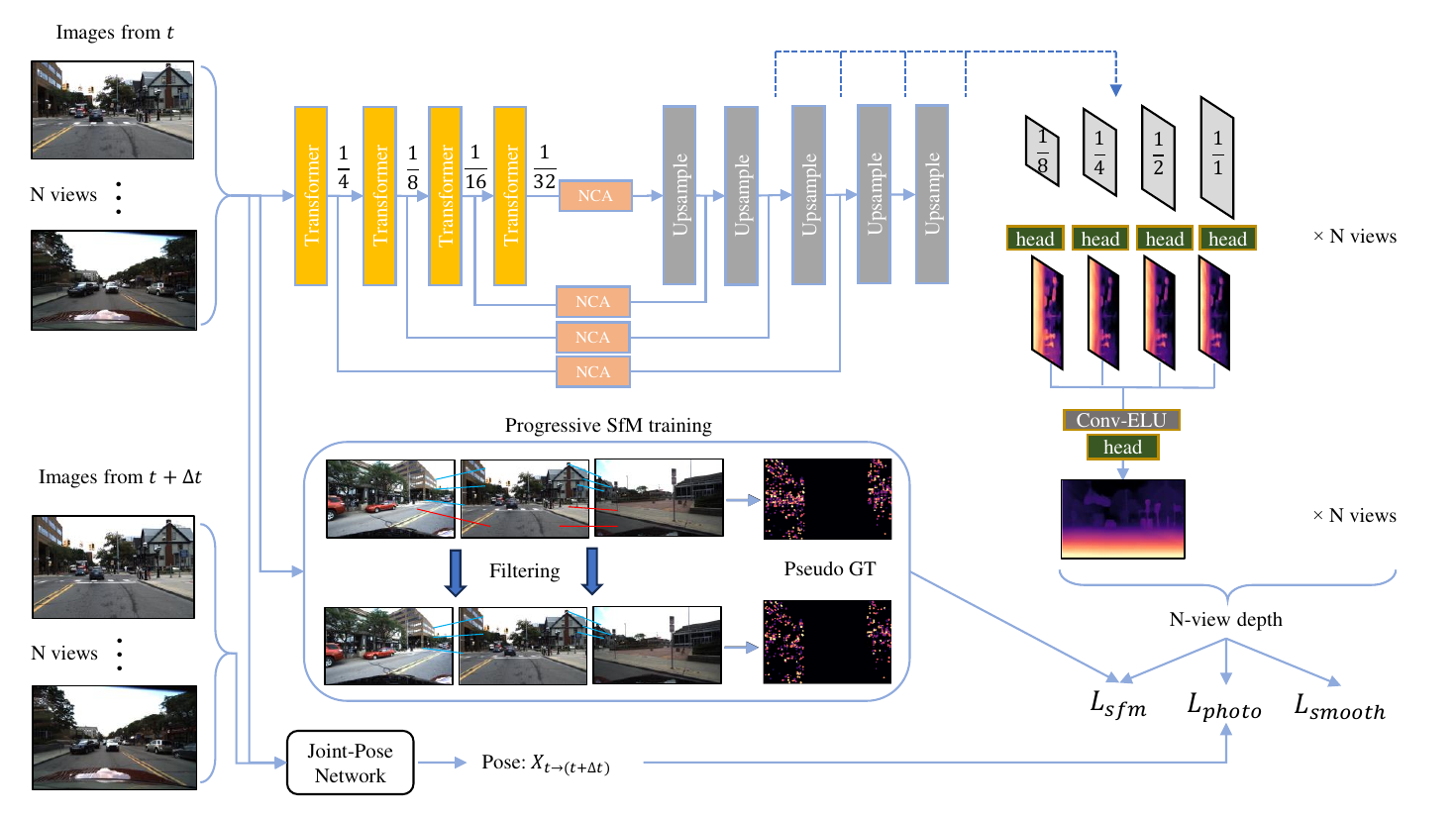}
\end{center}
\caption{Overview of our proposed FSM method. Images across camera views are concatenated and fed to a transformer-based depth network. Three losses are used for training this FSM depth network: (1)$L_{sfm}$ is calculated by using pseudo groundtruth generated from SfM pipeline. (2)$L_{photo}$ is calculated using spatial/temporal frames transformed by predicted pose changes $X_{t+\Delta t}$ and extrinsic matrices. (3) $L_{smooth}$ is calculated on individual depth maps that ensure edge-aware smoothness as commonly used in self-supervised monodepth methods\protect\cite{godard2017unsupervised,godard2019digging}.}
\label{fig:overview}
\end{figure*}
\subsection{Formulation}
Our definition of FSM follows previous works \cite{guizilini2022full,wei2022surrounddepth,xu2022multi}. In this approach, multiple cameras are placed around a vehicle to cover the 360-degree surroundings, with each camera having a small amount of overlapping area across views. Captured video sequences are used for training, with the frame captured by camera $n$ at time $t$ represented as $I_{t}^{n}$, where the camera ID $n \in \{1, 2, ..., N\}$.
The objective of FSM is to predict the corresponding depth for each of the cameras $D_{t}^{n}$.
In this paper, we assume that images from all views with the same timestamp should be processed jointly for cross-view context aggregation within the FSM depth network. Therefore, our method aims to learn a model F that takes in images from all camera views at time $t$ as input and outputs the depth map for each view: 
\begin{equation}
\begin{aligned}
  D_{t}^{1},D_{t}^{2}...D_{t}^{N}  = F(I_{t}^{1},I_{t}^{2}...I_{t}^{N})
\end{aligned}
\end{equation}
To learn depth without supervision, previous methods typically apply a photometric loss that minimizes the structural similarity \cite{wang2004image} (SSIM) between the target image $I_{tgt}$ and the synthesized target image $\hat{I}_{tgt}$, along with additional $L_1$ loss \cite{godard2017unsupervised,zhou2017unsupervised}:

\begin{equation} \footnotesize
\begin{aligned}\label{eq2}
  L_{photo} = & \frac{a(1-SSIM(I_{tgt}, \hat{I}_{tgt}))}{2} + (1-a) \lVert I_{tgt}- \hat{I}_{tgt} \rVert
\end{aligned}
\end{equation}

To obtain a synthesized target image, a pose network is needed to predict ego-motion of the camera frames. We follow \cite{wei2022surrounddepth} and apply the joint-pose network to predict one unified pose transformation. The pose network receives temporal adjacent frames $I_{t}^{n}$ and $I_{t+\Delta t}^{n}$, where $\Delta t \in \{-1,1\}$, from all camera views and predicts a rigid transform $X_{t \rightarrow (t+\Delta t)}$ that represents the movement of the cameras. 
The acquired pose transform is subsequently projected to every camera view using the extrinsic matrix $R^n$.This allows warping of neighboring frames to the target frame through:
\begin{equation}
\begin{aligned}\label{eq3}
  \hat{I}_t^n = proj(({R^n})^{-1}X_{t \rightarrow (t+\Delta t)}R^n, K^n , D_t^n, I_{t+\Delta t}^{n})
\end{aligned}
\end{equation}
$K^n$ denotes intrinsic matrix for n-th camera.
For spatially adjacent frames $I_{t}^{n}$, $I_{t}^{n+\Delta n}$ where $\Delta n \in \{-1,1\}$, Eq. \ref{eq3} is modified as: 
\begin{equation}
\begin{aligned}\label{eq4}
  \hat{I}_t^n = proj( (R^{n+\Delta n})^{-1}R^{n}, K^n , D_t^n, I_{t}^{n+\Delta n})
\end{aligned}
\end{equation}

The projection $proj$ is a set of fully differentiable operations including grid sampling and interpolation.
During training, the pose network and depth network are jointly learned by gradient update through Eq. \ref{eq2}.

As we have suggested earlier, the estimated depth from traditional self-supervised monodepth methods is scale-ambiguous. The final evaluation of these methods is done by applying median-scaling $\frac{median(D_{gt})}{median(D)}$ to the final depth predicted from the model against the groundtruth. 
With overlapping areas and extrinsic matrices, the FSM model can learn the real-world scale, and thus the evaluation does not require median-scaling. We refer to this as scale-aware evaluation, in contrast to scale-ambiguous evaluation.
In this paper, we focus on improving the accuracy of scale-aware evaluation as it is more real-world application friendly.

\subsection{Scale-Aware Full Surround Monodepth}
\subsubsection{Approach Overview.}

An overview of our method can be viewed in Fig. \ref{fig:overview}.  
We construct an encoder-decoder transformer-based network for FSM depth prediction. 
We stack images from multiple cameras into a tensor with shape (B, N, C, H, W), where each alphabet represents batch size, camera view, channel, height, and width. The encoder encodes the stacked tensor and outputs scaled features with height and width reduced to $\frac{1}{32}$ of the original resolution. The features are then restored to the original resolution by the decoder network through upsampling layers. 
Skip connections are built upon the 4 scales of encoded features, which are first processed by NCA modules before being merged with the decoded features.  
The final depth prediction is derived by integrating depth estimates from the depth network across various resolutions.

To construct a loss to train this depth network without supervision, we apply a joint-pose network to predict ego-motion of N views following \cite{wei2022surrounddepth}. The pose network receives raw images from N views at time $t$ and $t+\Delta t$, and predicts 6 numbers that encapsulate the axis-angles and translations in 3D space. 
The predicted pose and depth can thus be used to reconstruct the target frame from temporal adjacent frames using Eq. \ref{eq3}. Along with the synthesized target image from neighboring views, the final photometric loss $L_{photo}$ is calculated using Eq. \ref{eq2}.

We optimize an SfM supervision pipeline for training our depth network. This pipeline first conducts point matching among overlapping regions across views, and creates a sparse depth map based on triangulation and epipolar constraints that envelops approximated true scale. 
We adopt a two-round progressive training strategy to train the network. In the first round, we apply strong supervision from SfM with sufficient points for enforcing scale-awareness. In the second round, we only use filtered valid points for better alignment in different views. 
During inference, only the depth network is needed for the FSM depth prediction. 


\subsubsection{Depth Network Structure Optimization.}

\begin{figure}[t]
\begin{center}
\includegraphics[width=1.0\linewidth]{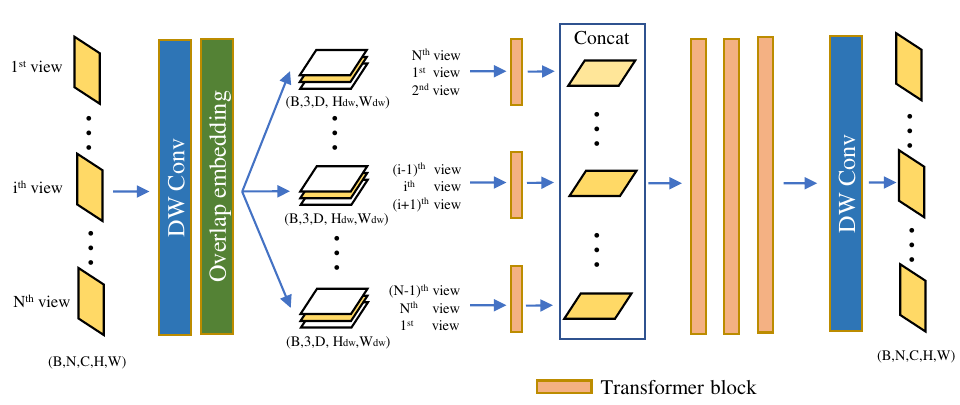}
\end{center}
   \caption{ Illustration of NCA. Tensor shape (B, N, C, H, W) follows the actual shape of the feature extracted from the encoder.}
\label{fig:SCVT}
\end{figure}



We categorize the context provided by surrounding views into two distinct types: neighbor context and global context. Neighbor context, derived from adjacent views, is intricately interrelated due to its significant spatial overlap. Consequently, we regard them as strongly correlated. 
In contrast, the global context offers a broader perspective within a single timestamp. 
We hypothesize that the depth model's scale-awareness will be amplified when the depth network effectively integrates both global and local contexts.

To address this, we propose a Neighbor-enhanced Cross-view Attention (NCA) module (see Fig. \ref{fig:SCVT}).
The NCA module is used to aggregate context information across camera views globally and within strongly correlated adjacent views. 
The NCA involves several steps: (1) Downsampling and embedding. (2) Applying transformer blocks to representations of adjacent views. (3) Applying transformer blocks to global representations. (4) Upsampling. Details follow.

The input of the NCA module are the features with shape (B, N, C, H, W) extracted from the encoder.
These features are first processed by depth separable convolution \cite{chollet2017xception} and downsampled to (B, N, C, H$_{dw}$, W$_{dw}$) for efficiency considerations. H$_{dw}$ and W$_{dw}$ equals to $\frac{1}{32}$ of the original input size.
We use the overlapped embedding scheme as a contrast to the traditional position embedding, aiming to preserve spatial continuity within each camera view. Specifically, we apply an independent embedding for each camera view in the starter of the NCA module, achieved by a convolution layer (kernal\_size=3, stride=2, padding=1) and normalization layer, applied to the N-view features separately. 

The embedded features are then fed to the attention blocks for further processing.
Here, we utilize two types of inputs with different feature representations. One focuses only on neighbor context exchange, and the other focuses on global context exchange within all camera views. To achieve this, we first group the embedded center features and neighbor-view feature to form new sets of embedded representations, such as $X^{n-2}X^{n-1}X^{n}, X^{n-1}X^{n}X^{n+1}, X^{n}X^{n+1}X^{n+2}, \dots$ (See left part of Fig. \ref{fig:SCVT}).
We then incorporate individual transformer blocks to each item in the set. However, since this operation is computationally expensive in a multi-camera setting, we only apply a single multi-head attention block for each grouped feature in this stage. 

For global context exchange, we regroup the features by extracting features from the previously processed sets and only choose the center view features $X^{n-1}, X^{n}, X^{n+1}, \dots$. The neighbor-view features in one set are then discarded since the purpose of them is to provide context info to the center view (marked as 'white tensor' in Fig. \ref{fig:SCVT}). 
The center-view features are concatenated together and then fed to the multi-head attention blocks. In practice, we first conduct neighbor-view context exchange and then global context exchange.
The final output from NCA module is extracted by applying a depth-wise conv to restore the features back to original size.  

The transformer blocks of NCA is implemented based on Segformer block \cite{xie2021segformer} which is composed of a standard attention module and Mix-FFN. The Mix-FFN is implemented with a combination of MLP-Conv2d-GELU-MLP. 
For the Conv2d and MLP layers inside the Mix-FFN, we treat the camera channel (N) and hidden layer dimension (dim) as equal with channel manipulation: (B, N $\times$ dim, H$_{dw}$, W$_{dw}$). After Mix-FFN, the features are restored back to the 5D tensor and features from different views are separated again. 




We further utilize Segformer as encoder of the depth network to adapt the NCA module.
During forward, camera from each view is treated the same as individual frame during the encoding process.
For the decoder of depth network, we build several upsampling blocks to gradually restore the feature resolution. Each upsampling block consists of interpolation and convolution operations.
The NCA modules are positioned at four different resolutions, serving as a bridge between the encoder and decoder.


The final depth prediction is generated by gathering predicted depth map from various resolutions. 
High-resolution depth maps typically cover more high-frequency details but may show lower structural consistency, while low-resolution depth maps exhibit the opposite behavior \cite{miangoleh2021boosting}. 
To combine the advantages of both high and low-resolution depth maps, we first predict depth map for each scale by passing the features through prediction heads. The prediction heads are composed of Conv2D blocks and a sigmoid function ("head" in Fig. \ref{fig:overview}). we then unify the size of the depth maps to match the input size. 
Next, we concatenate the predicted depths and apply a combination of a 3x3 Conv-ELU block, followed by another prediction head, to retrieve the final depth estimation.
During training, we apply a weighted training loss for all four resolution depths ($\frac{1}{8}$, $\frac{1}{4}$, $\frac{1}{2}$ and the final output $\frac{1}{1}$ depth), with greater weight given to the final depth map processed by the aggregation module, reflecting the fact that the lower-resolution depth maps only serve as supplements to the high-resolution depth map.

\subsubsection{Training Pipeline Optimization.}

To initialize a real-world scale, we apply structure-from-motion (SfM) pipeline to provide weak supervision signal for FSM depth learning. 
SfM can quickly guide the network to learn true scale compared to relying only on spatial constraints in loss term \cite{wei2022surrounddepth}. 
Here, we illustrate how we further optimize the SfM training pipeline. 

We design a two-round progressive training scheme. In the first round, we use Superglue \cite{sarlin2020superglue} as an alternative to traditional SIFT descriptors for better feature matching. 
With known camera matrices and triangulation, we can create a pseudo groundtruth map that contains depth values that are close to the real-world scale. The pseudo groundtruth is then used as a supervision signal to train the network. In this phase, all the matching points contribute to SfM training, which help learn real-world scale.

The first-round training derives a good initialization for scale-awareness. In the second round, we use only valid points as constraints to further facilitate the network training. This also helps reduce the negative impact of incorrectly matched points. To this end, we rank the loss calculated in each matching point, remove top 33\% points with a large loss value, and keep the rest for $L_{sfm}$ calculation.
Our final training loss can be written as a combination of SfM loss $L_{sfm}$, spatial-temporal photometric loss $L_{photo}$ and smoothness term $L_{smooth}$:
\begin{equation}
\begin{aligned}
  L_{h,w}^{1} &= \sigma_1 L_{sfm}(P)+ \sigma_2 L_{photo} + L_{smooth}, \\
  L_{h,w}^{2} &= \sigma_1 L_{sfm}(\hat{P}) + \sigma_2 L_{photo} + L_{smooth}, \\
  L_{final} &=  p_1 L_{h,w} + p_2 L_{h/4,w/4} \\
  &+ p_2 L_{h/8,w/8} + p_2 L_{h/16,w/16} 
\end{aligned}
\end{equation}
Where $P$ means original matching points and $\hat{P}$ means filtered matching points. In practice, we set $\sigma_1=0.1$ in the first round, and set $\sigma_1=0.005$ in the second round to allow more self-adjustment from self-learned spatial constraints. $\sigma_2=0.5$, $p_1=\frac{1}{2}$ and $p_2=\frac{1}{6}$ for both training rounds.

\section{Experiments}

\begin{table*}
\footnotesize
\begin{center}
\begin{tabular}{|c|l|c|c|c|c|c|c|c|}
\hline
 & Method & $abs\_rel$ & $sq\_rel$ & $rmse$ & $rmse\_log$ & $\delta<1.25 $ & $\delta<1.25^2$ & $\delta<1.25^3$ \\
\hline\hline
\multirow{4}{*}{Median-scale}& Monodepth2 \cite{godard2019digging}  & 0.362  & 14.404  & 14.178  & 0.412  & 0.683  & 0.859  & 0.922 \\
& FSM* \cite{kim2022self}  & 0.219  & 4.161  & 13.163  & 0.327  & 0.703  & 0.880  & 0.940 \\
& VFF \cite{kim2022self} & 0.221  &  3.549 &  13.031 &  0.323 &  0.681 &  0.874 &  0.940 \\
& Surrounddepth \cite{wei2022surrounddepth} &  0.200  & 3.392    & 12.270   & 0.301 & 0.740   &   0.894  &  0.947  \\


& Ours  &   0.189  &   3.130  &  12.345  &   0.299  &   0.744  &   0.897  &   0.949    \\
\hline
\multirow{4}{*}{Scale-aware}
& FSM* \cite{kim2022self}   & 0.228  & 4.409  & 13.433  & 0.342  & 0.687  & 0.870  & 0.932 \\
& VFF\cite{kim2022self} & 0.218  & 3.660  &  13.327 &  0.339  &  0.674  &  0.862  &  0.932 \\
& Surrounddepth \cite{wei2022surrounddepth} & 0.208  &   3.371 &   12.977  &   0.330  &   0.693  &   0.871  &   0.934 \\
& SA-FSM (Ours)  &   \textbf{0.187}  &  \textbf{3.093}  &  \textbf{12.578}  &   \textbf{0.311}  &  \textbf{0.731}  &  \textbf{0.891}  &   \textbf{0.945} \\
\hline

\end{tabular}
\end{center}
\caption{Results overview on DDAD dataset. Rows with "Median-scale" indicates the evaluation is conducted with the auxiliary median-scaling step, while "Scale-aware" rows directly evaluate the predicted depth against the groundtruth. We use the same model for evaluation in both ways. FSM$^*$ indicates reproduced results from VFF\protect\cite{kim2022self}. 
}
\label{tab:DDAD}
\end{table*}

\begin{table*}
\footnotesize
\begin{center}
\begin{tabular}{|c|l|c|c|c|c|c|c|c|}
\hline
 & Method & $abs\_rel$ & $sq\_rel$ & $rmse$ & $rmse\_log$ & $\delta<1.25 $ & $\delta<1.25^2$ & $\delta<1.25^3$ \\
\hline\hline
\multirow{4}{*}{Median-scale}& Monodepth2 \cite{godard2019digging}  & 0.287  & 3.349  & 7.184  & 0.345  & 0.641  & 0.845  & 0.925 \\
& FSM* \cite{kim2022self}  & 0.301  & 6.180  & 7.892  & 0.366  & 0.729  & 0.876  & 0.933 \\
& VFF \cite{kim2022self} & 0.271  &  4.496 &   7.391 &  0.346 & 0.726 &  0.879 &   0.934 \\
& Surrounddepth \cite{wei2022surrounddepth} & 0.245  &  3.067 & 6.835  & 0.321 &  0.719   &  0.878  &  0.935  \\
& Ours  &  0.245  &  3.454  &  6.999  &  0.325  &  0.725  &  0.875  &  0.934  \\
\hline
\multirow{4}{*}{Scale-aware}
& FSM* \cite{kim2022self}  & 0.319  & 7.534  & 7.860  & 0.362  & \textbf{0.716}  & 0.874  & 0.931 \\
& VFF \cite{kim2022self}  & 0.289   & 5.718  &  7.551 &   \textbf{0.348}  &   0.709  &  \textbf{0.876}  &  \textbf{0.932} \\
& Surrounddepth \cite{wei2022surrounddepth}  &  0.280  &  \textbf{4.401} &  7.467  &  0.364  &  0.661   &  0.844   &  0.917  \\
& SA-FSM (Ours)  &  \textbf{0.272}  &  4.706  &  \textbf{7.391}  &  0.355  &  0.689  &  0.868  &  0.929  \\
\hline

\end{tabular}
\end{center}
\caption{Results overview on nuScenes dataset. Same specifications as for DDAD.}
\label{tab:nuScenes}
\end{table*}

\subsection{Experimental Settings}
\subsubsection{Datasets:}
We conduct our experiments on two public datasets: DDAD \cite{guizilini20203d} and nuScenes \cite{caesar2020nuscenes}. Both datasets have 6 cameras attached to a vehicle to capture surround-view sequences. 
DDAD collects data in urban setting that covers cities in the US and Japan. It contains 12650 training samples, each sample contains frames from 6 views. For validation, there are 3950 samples with groundtruth depth available.
NuScenes data is collected in Boston and Singapore, it contains 1000 sequences of videos. 700 of them are for training, 150 for validation, and 150 for testing.

\subsubsection{Implementation details:} We train our model with 4 $\times$ 32G GPUs. We used the Adam optimizer throughout the training process. The initial learning rate for first round training was $6e^{-5}$, and $5e^{-5}$ for the second round. 
The input size of DDAD is 384 $\times$ 640. Following previous works\cite{wei2022surrounddepth,guizilini2022full}, we only validate the depth within 200m. We follow Wei \cite{wei2022surrounddepth} in preparing datasets for training and validation, and we apply the same occlusion masks for the occlusion parts which came from the data collection vehicle itself during training and evaluation. The input size for nuScenes is 352 $\times$ 640, and static frames from the sequence are filtered. 
We apply the same hyper-parameters for both nuScenes and DDAD datasets. 
Several evaluation metrics are applied including absolute relative error (abs\_rel), squared relative error (sq\_rel), root mean square error (rmse), log variant of root mean square error (rmse\_log), and pixel depth with error less than a threshold ($\delta$) are used for evaluation. The first four evaluation metrics are lower the better, while for $\delta$, the bigger the better.

\subsection{Comparison with State-of-the-Art Methods}

\subsubsection{Results on DDAD:}
Table \ref{tab:DDAD} presents a comparison of the accuracy of the DDAD dataset with previous FSM methods. Our proposed method achieves SOTA accuracy on all evaluation metrics for scale-aware depth prediction and exhibits a significant improvement in accuracy compared to the second-best method \cite{wei2022surrounddepth}, which shares a similar SfM pipeline and joint-pose network. 
The overall performance gap against other methods is shortened median-scale evaluation, but our method still achieved a lowest $abs\_rel$ score. 
Considering both evaluation ways, we found that the performance of our method under both median-scale and scale-aware evaluation was similar, which suggests that our method is practical and can be used without getting true scale from groundtruth.

\subsubsection{Results on nuScenes:}
Table \ref{tab:nuScenes} shows the accuracy comparison on nuScenes with previous FSM methods. Without changing hyper-parameters, our method reached better performance than others over $abs\_rel$ and $rmse$. And our method outperforms Surrounddepth \cite{wei2022surrounddepth} in multiple criteria, which suggests our SfM pipeline and depth network are superior.  
In general, our model performs better in DDAD than nuScenes.
We believe the reason for this difference in performance is due to the fact that nuScenes has smaller overlapping areas between camera views compared to DDAD. As our method relies on the correspondence on overlapping areas, such as neighboring context aggregation, the performance is inevitably affected by the amount of overlapping areas between views.

\subsection{Ablation Study}

\begin{figure*}
\centering
\includegraphics[width=0.85\textwidth]{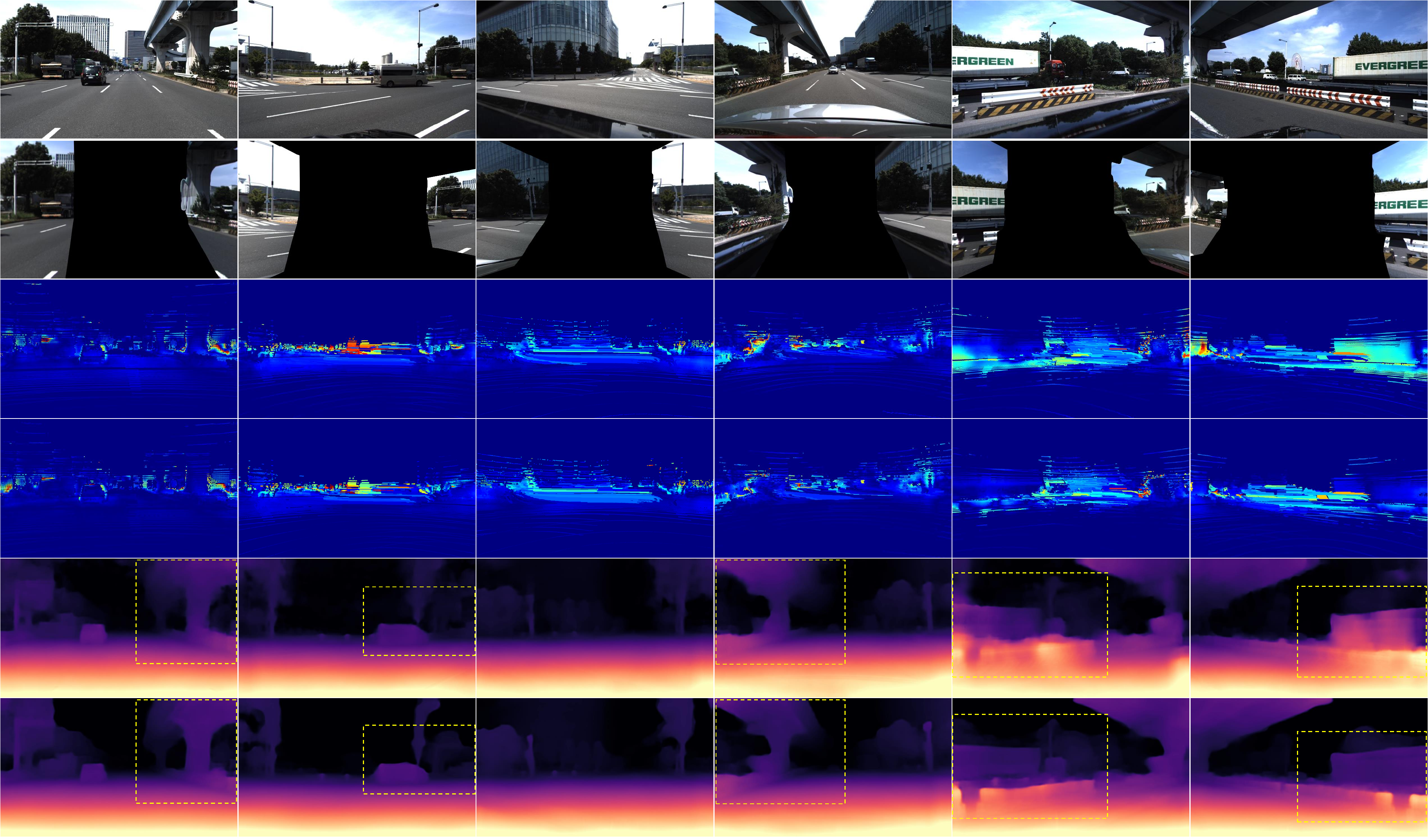}
   \caption{Visualization results. First row: RGB images from six views in DDAD. Second row: Images warped from adjacent views. Third row: $Abs\_rel$ error map predicted by previous SOTA Surrounddepth \protect\cite{wei2022surrounddepth}. Forth row: $Abs\_rel$ error map predicted by our method, color close to the background blue indicates a better accuracy in groundtruth pixel location. Fifth row: Depth predicted by Surrounddepth. Sixth row: Depth predicted by our model. Areas that shows depth improvements are highlighted with yellow rectangles. }
\label{fig:depth_vis}
\end{figure*}

\begin{table}
\small
\centering
\resizebox{0.47\textwidth }{!}{
\begin{tabular}{|c|c|c|c|c|c|}
\hline
Structure opt. & Training opt. & $abs\_rel$ & $sq\_rel$ & $rmse$ & $\delta<1.25$ \\
\hline\hline
\xmark &  \xmark & 0.208  & 3.371 & 12.977 & 0.693 \\
\xmark &  \cmark & 0.212 &  3.493 &  13.509  & 0.674 \\
\cmark &  \xmark & 0.201  & 3.240  & 12.943  & 0.700  \\
\cmark &  \cmark & \textbf{0.187}  & \textbf{3.093} & \textbf{12.578} & \textbf{0.731} \\
\hline
\end{tabular}
}
\caption{Ablation study on the proposed depth network structure and training pipeline optimization.} 
\label{tab:ablation_1}
\end{table}

\begin{table}
\small
\centering
\resizebox{0.47\textwidth }{!}{
\begin{tabular}{|c|c|c|c|c|c|}
\hline
Structure opt. & Training opt. & $abs\_rel$ & $sq\_rel$ & $rmse$ & $\delta<1.25$ \\
\hline\hline
CVT &  \xmark & 0.206  & 3.280 & 12.610 & 0.699\\
NCA &  \xmark & 0.201  & 3.240 & 12.943 & 0.700  \\
CVT &  \cmark & 0.191  & \textbf{3.091} & 12.588 & 0.706  \\
NCA &  \cmark & \textbf{0.187}  & 3.093 & \textbf{12.578} & \textbf{0.731} \\
\hline
\end{tabular}
}
\caption{Ablation study on CVT vs. NCA.} 
\label{tab:ablation_2}
\end{table}

\begin{table}
\small
\centering
\resizebox{0.47\textwidth }{!}{
\begin{tabular}{|c|c|c|c|c|c|}
\hline
Descriptor & Progressive Training & $abs\_rel$ & $sq\_rel$ & $rmse$ & $\delta<1.25$ \\
\hline\hline
SIFT & \xmark & 0.201 &  3.240 &  12.943 &  0.700\\
SIFT &  \cmark & 0.194 & 3.148 & 12.920 & 0.708\\
SG &  \xmark & 0.195  & 3.106  &  12.775  &  0.711 \\
SG &  \cmark & \textbf{0.187}  & \textbf{3.093} & \textbf{12.578} & \textbf{0.731} \\
\hline
\end{tabular}
}
\caption{Ablation study on progressive training. Depth network structure optimization is enabled in these ablation experiments.} 
\label{tab:ablation_3}
\end{table}

This section analyzes the effectiveness of our proposed network structure and training pipeline optimization.

\subsubsection{Effect of Depth Network Structure Optimization:} 

In Table \ref{tab:ablation_1}, we set up controlled experiments, giving the option of whether to apply structure optimization, training optimization, or both. The results show that, with or without training optimization, the application of depth network structure optimization leads to a large gain compared to scenarios without structure optimization. The depth estimation reaches the highest accuracy in terms of all metrics when both structure optimization and training optimization are applied.

\subsubsection{Effect of Training Pipeline Optimization:} 

In Table \ref{tab:ablation_1}, with structure optimization enabled, the application of training pipeline optimization can further improve the results. We also observe that when using training optimization without structure optimization, there is a negative effect on accuracy. This phenomenon can be attributed to the more complex training pipeline that comes with the use of Superglue and progressive training.
The network structure thus faces more strict requirements for learning and adaptation. The use of a transformer-based structure in the model mitigates this negative effect, turning it into a positive outcome.



\subsubsection{CVT vs. NCA:} 
We replace the NCA module in our depth network with the CVT module from Surrounddepth, and provide a comparison in Table \ref{tab:ablation_2}. The main difference between two designs lies in that CVT only considers global context information while our NCA module incorporates both neighboring and global context.
The results demonstrate that NCA consistently outperforms CVT in both scenarios, with or without training optimization, and exhibits improvements in specific measures.
Without training optimization, switching from CVT to NCA improve the abs\_rel from 0.206 to 0.201 and sq\_rel from 3.280 to 3.240. 
With training optimization, the use of NCA improves the abs\_rel score from 0.191 to 0.187 and it improves $\delta$ to 0.730 with a large margin of 0.025 compared with CVT setup.
In addition, our NCA model requires fewer FLOPs (34G) in comparison to CVT (55G). This is notable, especially considering that both CVT and NCA were developed with comparable computational resources, and yet NCA consistently demonstrates superior performance.

\subsubsection{Effect of Progressive Training:} 
Table \ref{tab:ablation_3} provides ablation study on the progressive training scheme by setting descriptor options (SIFT and SuperGlue). We observe that progressive training improves the result for both types of descriptors, which validates its effectiveness. With SuperGlue as descriptor, the progressive training scheme reaches the overall best accuracy.

\subsubsection{Efficiency Analysis:} 
We compare the computation complexity and inference speed with the prior state-of-the-art Surrounddepth method. Our model utilizes fewer computational resources: 361G FLOPs (Ours) vs. 406G FLOPs (Surrounddepth). Our depth prediction for 6 cameras operates at 60 ms per timestamp, compared to 100 ms from SurroundDepth, indicating a 1.67x speed increase. This indicates the efficient nature of our approach in achieving faster and more economical depth predictions for full surround depth.

\subsubsection{Depth Visualization:} An overview of depth prediction results is shown in Fig. \ref{fig:depth_vis}. From the second row, we can see that the warped image result matches the original RGB image, which indicates well learned true scale. The third and fourth rows show the error maps, calculated by the absolute difference with $\frac{|D - D_{gt}|}{D_{gt}}$. Warmer colors indicate larger errors, while cooler colors represent smaller errors. 
Our results show that the errors on the vehicle body and the ground area are lower compared to other methods. Moreover, the overall large error area is largely reduced.
In the fifth and sixth rows, we provide an illustration of the predicted depth maps from Surrounddepth \cite{wei2022surrounddepth} and ours, respectively. Our method produces clearer edges, smoother ground surfaces, and accurate depth.

\section{Conclusion}


We propose to enhance scale-awareness from two aspects of depth network structure optimization and training pipeline optimization. First, we construct a transformer-based depth network with NCA modules to aggregate cross-view context in both neighboring and global views. Second, we formulate a transformer-based feature matching scheme with progressive training for better alignment between different views. Our SA-FSM method can improve scale-aware depth prediction and achieve superior performance over previous FSM works on the DDAD benchmark.


\bibliographystyle{named}
\bibliography{ijcai24}

\end{document}


\maketitle

\section{Overview}

In this supplementary material, we present more algorithm details and experimental results.
\begin{itemize}
    \item We conduct additional ablation studies to examine the hyperparameters used in network structure and training pipeline optimization.
    \item We provide the network architecture and more implementation details of our method.
    \item We present more qualitative results on DDAD and nuScenes.
\end{itemize}

\section{More Ablation Studies}

\subsubsection{NCA Analysis.}

We conduct an analysis of the effectiveness of our proposed Neighbor-enhanced Cross-view Attention (NCA) module on the DDAD dataset in Table \ref{tab:NCA_ablation}. To explore the effect of the number of transformer blocks in the NCA module, we experimented with three schemes (i.e., depth=1/3/5). The results indicate that depth=3 yields the best overall performance, e.g., with $abs\_rel$ reduced to 0.187.
Under the same setting of depth=3, we conduct an experiment by removing the neighbor enhancement in the NCA module. The results show that the performance drops in terms of all the metrics without neighbor context exchange (row 2 vs. row 4 in Table \ref{tab:NCA_ablation}).


\subsubsection{Training pipeline optimization.}
\begin{figure}[tbh]
\begin{center}
\includegraphics[width=0.95\linewidth]{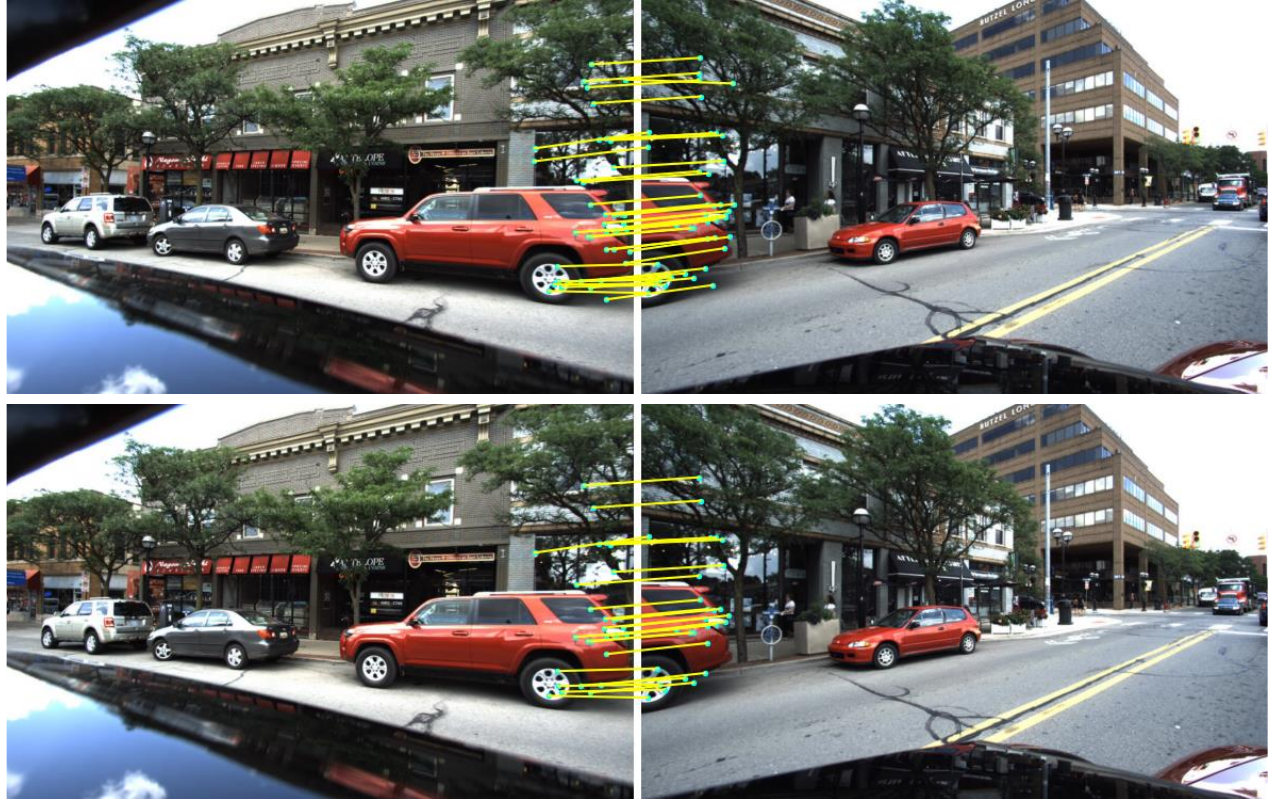}
\end{center}
   \caption{Visualization of the matched points used in the proposed progressive training. Matching points are used in the first round and filtered valid points are used in the second round.}
\label{fig:superglue_vs_sift}
\end{figure}

Our training pipeline applies progressive training. 
We conduct experiments and set different filtering ratios for the second training round \{100\%, 66\%, 33\%, 0\%\}. Particularly, 0\% means that all the matched points are used in the second training round without filtering, and 100\% means that $L_{sfm}$ is disabled in the second training round.
Using both the transformer-based depth network and training pipeline optimization, the abs\_rel results for these ratios are {0.194, 0.191, 0.187, 0.191}. We thus select 33\% for in the final setup. Moreover, we provide a visualized comparison of matched points used in the two training rounds in Fig. \ref{fig:superglue_vs_sift}. Matched points with greater loss are filtered in the second round to facilitate the network training. 


\section{Implementation Details}
Our model is trained on 4 $\times$ 32G GPUs using PyTorch. It takes 43 hours to complete two rounds of training. The batch size is specified as 6\footnote{There are 6 frames from different camera views in one timestamp.} and the input size is $6 \times 3 \times 384 \times 640$. We employed the Adam optimizer with $\beta{1}$ = 0.9 and $\beta{2}$ = 0.999. To compute the photometric reprojection loss in the loss function:
\begin{equation} \footnotesize
\begin{aligned}\label{eq2}
  L_{photo} = & \frac{a(1-SSIM(I_{tgt}, \hat{I}_{tgt}))}{2} + (1-a) \lVert I_{tgt}- \hat{I}_{tgt} \rVert
\end{aligned}
\end{equation}
we assigned SSIM weights of $\alpha$ = 0.85. 


\section{Network Architecture}
We utilize an encoder-decoder transformer-based depth network and a joint-pose network, which are detailed in Table \ref{tab:depthnet} and Table \ref{tab:posenet}, respectively. The symbol $\oplus$ denotes feature concatenation. The predicted depth maps take the form of N × 1 × H × W tensors, while the predicted poses consist of two 6-dimensional vectors representing translation. The output dimension of the last row in Table \ref{tab:posenet} is 1 × 2 × 1 × 6 which is redundant, and the actual final vector is 1 × 1 × 1 × 6.

\begin{table}[!t]
\footnotesize
\begin{center}
\resizebox{0.46\textwidth }{!}{
\begin{tabular}{|c|c|c|c|c|c|}
\hline
SA-FSM (ours)  & $abs\_rel$ & $sq\_rel$ & $rmse$ & $\delta<1.25$ \\
\hline\hline

NCA depth=1  & 0.188 & 3.057 & 12.620 & 0.727 \\
\hline
NCA depth=3  & \textbf{0.187} & 3.093 & \textbf{12.578} & \textbf{0.731} \\
\hline
NCA depth=5  & 0.191 & \textbf{3.019} & 12.767 & 0.718 \\
\hline
NCA depth=3 w/o neighbor  & 0.193 & 3.150 & 12.615 & 0.725 \\
\hline
\end{tabular}
}
\end{center}
\caption{Ablation study on NCA. ``depth" indicates the number of transformer blocks in NCA for global context exchange. ``w/o neighbor" indicates NCA without neighbor context exchange. 
} 
\label{tab:NCA_ablation}
\end{table}

\begin{table}[!h]
\begin{center}
\includegraphics[width=0.98\linewidth]
{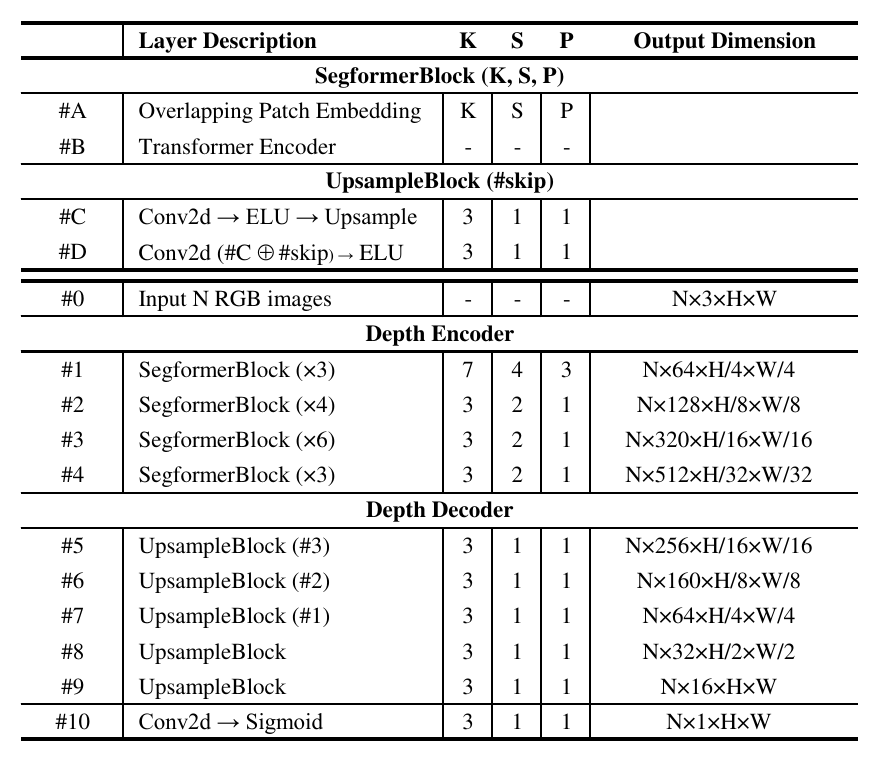}
\end{center}
\caption{Depth network structure. K, S and P indicate the kernel, stride and padding for convolution layer respectively. The first column contains IDs for distinguishing between layers. ``\#skip" indicates the skip connection of features between encoder and decoder. 
``\#C'' in Conv2d and ``\#3/\#2/\#1" in UpsampleBlock indicate jump connections originated from these layers.
}
\label{tab:depthnet}
\end{table}

\begin{table}[!h]
\begin{center}
\includegraphics[width=0.98\linewidth]
{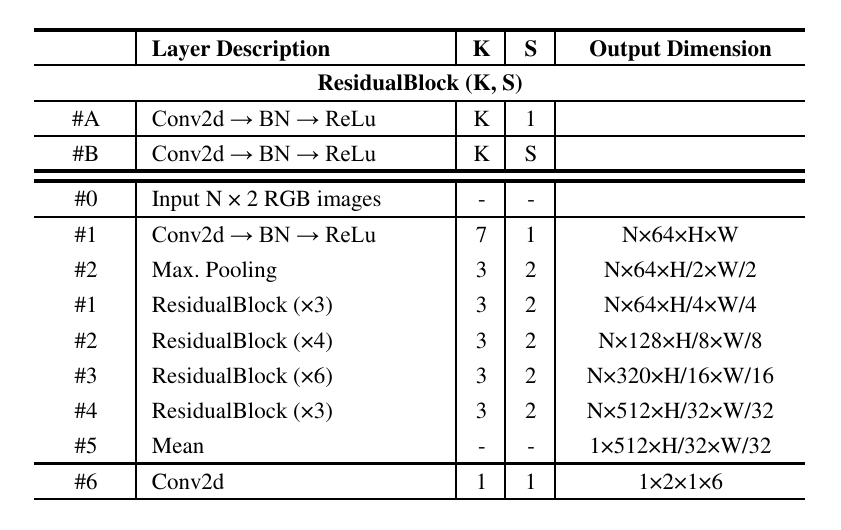}
\end{center}
\caption{Pose network structure. K and S indicate the kernel size and stride of the convolution layer respectively.}
\label{tab:posenet}
\end{table}




\section{More Qualitative Results}
\begin{figure*}[!h]
\begin{center}
\includegraphics[width=0.95\textwidth]{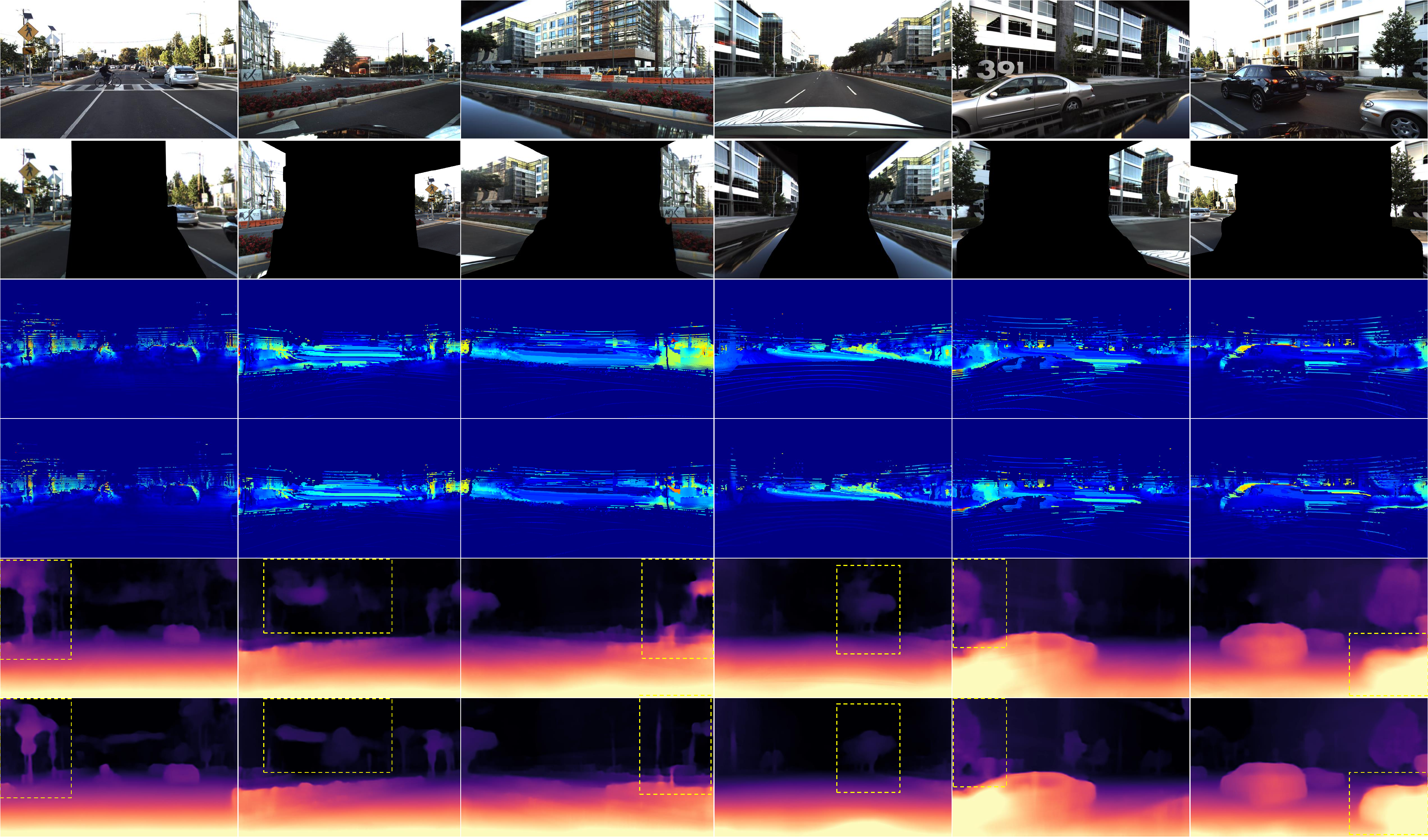}
\end{center}
   \caption{Visualization results on the DDAD dataset.}
\label{fig:visual_ddad}
\end{figure*}

\begin{figure*}[!h]
\begin{center}
\includegraphics[width=0.95\textwidth]{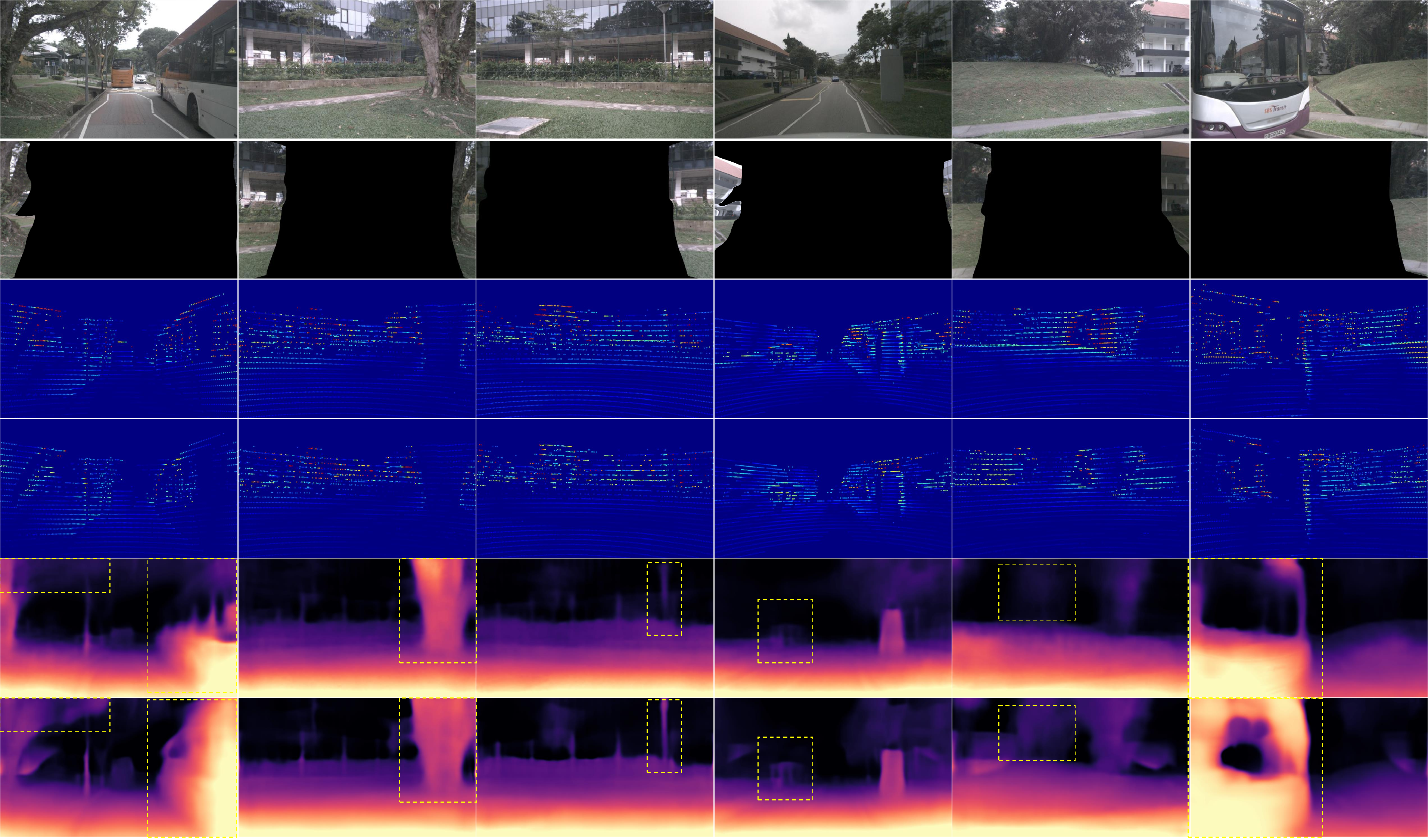}
\end{center}
   \caption{Visualization results on the nuScenes dataset. }
\label{fig:visual_nusc}
\end{figure*}

We provide more qualitative results on DDAD in Figure \ref{fig:visual_ddad} and on nuScenes in Figure \ref{fig:visual_nusc}. The first row shows RGB images from 6 views in the dataset. The second row shows images warped from adjacent views. The third and fourth rows display the respective $abs\_rel$ error map generated by Surrounddepth and our proposed model respectively, where the color close to the background blue indicates better accuracy in the ground truth pixel location. The fifth and sixth rows present the predicted depth maps of Surrounddepth and our method respectively, where yellow rectangles are used to highlight areas that exhibit improvements in depth accuracy. These visualization results demonstrate that our model can more accurately capture depth information and preserve clearer object edges in the scene.

